\documentclass{article}

\usepackage{LIreduced}

\usepackage{titlepic}

\usepackage[utf8]{inputenc}
\usepackage{graphicx}
\usepackage{amsfonts}
\usepackage{color}
\usepackage{amsthm}
\usepackage{amssymb}
\usepackage{enumitem}
\usepackage[mathscr]{euscript}
 \let\mathscr\relax
\usepackage[scr]{rsfso}
\usepackage{comment}
\usepackage{amsmath}

\usepackage[comma]{natbib}           
\title{Something Old, Something New: Grammar-based CCG Parsing with Transformer Models}

\author{Stephen Clark\\ Cambridge Quantum Computing\\ Oxford/Cambridge, UK}

\titlepic{\includegraphics[width=5cm]{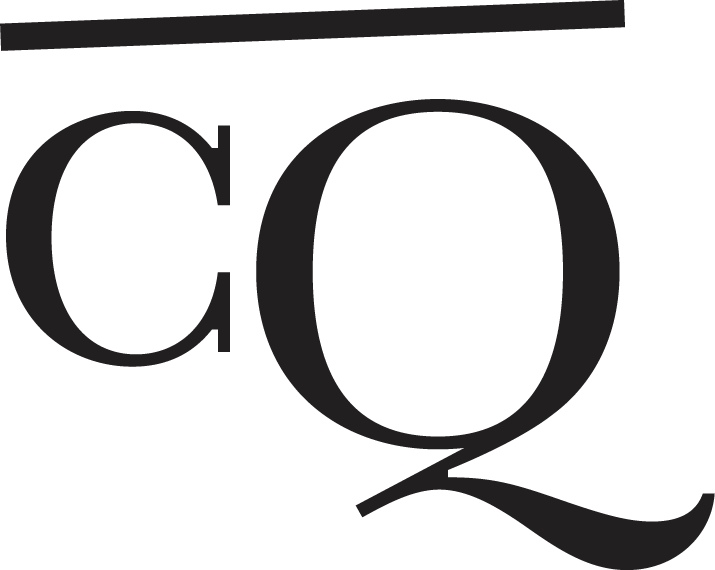}}

\date{September 2021}

\newcommand{\cf}[1]{\mbox{$\it{#1}$}}   
\newcommand{\reln}[1]{\mbox{\small{\it #1}\/}}
\newcommand{\head}[1]{_{\mbox{\tiny{\it #1}}}}
\newcommand{\feat}[1]{[\mbox{\footnotesize{\it #1}\/}]}

\begin{document}

\maketitle

\begin{abstract}
This report describes the parsing problem for Combinatory Categorial Grammar (CCG), showing how a combination of Transformer-based neural models and a symbolic CCG grammar can lead to substantial gains over existing approaches. The report also documents a 20-year research program, showing how NLP methods have evolved over this time. The staggering accuracy improvements provided by neural models for CCG parsing can be seen as a reflection of the improvements seen in NLP more generally. The report provides a minimal introduction to CCG and CCG parsing, with many pointers to the relevant literature. It then describes the CCG supertagging problem, and some recent work from \cite{tian-etal-2020-supertagging} which applies Transformer-based models to supertagging with great effect. I use this existing model to develop a CCG multitagger, which can serve as a front-end to an existing CCG parser. Simply using this new multitagger provides substantial gains in parsing accuracy. I then show how a Transformer-based  model from the parsing literature can be combined with the grammar-based CCG parser, setting a new state-of-the-art for the CCGbank parsing task of almost 93\% F-score for labelled dependencies, with complete sentence accuracies of over 50\%.
\end{abstract}

\section{Introduction}

Combinatory Categorial Grammar (CCG) is a lexicalized grammar formalism in the type-driven tradition, building on historical work by \cite{ajdukiewicz:35} and \cite{bar-hillel:53}. The original formalism, often referred to as classical categorial grammar, uses the rules of forward and backward application to combine the categorial types. CCG, developed over many years by Mark Steedman \citep{steedman:2000}, uses a number of additional combinatory rules to deal with ``movement" phenomena in natural languages -- syntactic environments in which phrases are moved from their canonical argument positions, often creating an unbounded dependency between the argument and predicate. Examples in English include questions and relative clause extraction \citep{rimell-etal-2009-unbounded}. This movement phenomena is what motivated Chomsky to develop transformational grammar \citep{chomsky:65}. Unlike transformational grammar, however, CCG is a ``monostratal" theory in which the apparent movement of syntactic units is handled by a single level of representation. Other approaches to categorial grammar include the type-logical approach \citep{moortgat:97}, in which linguistic types are the formulas of a logic and derivations are proofs, and the algebraic approach of the later work of Lambek \citep{lambek:08}, in which linguistic types are the partially-ordered objects of an algebra (specifically a \emph{pregroup}), and derivations are given by the partial order.

\begin{figure}
\begin{center}
     \includegraphics[width=12.5cm]{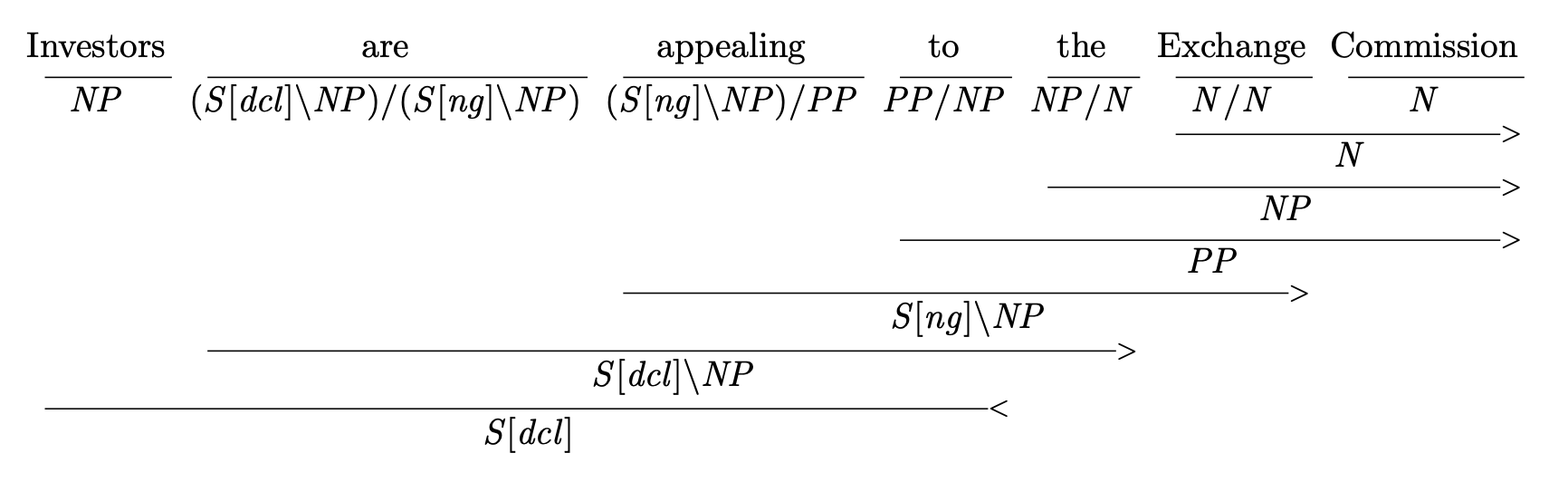}
\end{center}
\caption{Example derivation using forward and backward application.}
\label{fig:application_deriv}
\end{figure}

Figure~\ref{fig:application_deriv} gives an example CCG derivation using only the basic rules of \emph{forward} ($>$) and \emph{backward} ($<$) \emph{application}. The categorial types that are assigned to words at the leaves of the derivation are referred to as \emph{lexical categories}.\footnote{Figure~\ref{fig:application_deriv} follows the typical presentation for a CCG derivation with the leaves at the top.} The internal structure of categories is built recursively from atomic categories and slashes (`\textbackslash', `/') which indicate the directions of arguments. In a typical CCG grammar there are only a small number of atomic categories, such as \cf{S} for sentence, \cf{N} for noun, \cf{NP} for noun phrase, and \cf{PP} for prepositional phrase.\footnote{ The square brackets on some \cf{S} nodes in the example denote grammatical features such as \cf{[dcl]} for declarative sentence \citep{hockenmaier-steedman-2007-ccgbank}.} However, the recursive combination of categories and slashes can lead to a large number of categories; for example, the grammar used in the parsing experiments below has around 1,300 lexical categories. CCG is referred to as \emph{lexicalised} because most of the grammatical information---which is language-dependent and encoded in the lexical categories---resides in the lexicon, with the remainder of the grammar being provided by a small number of combinatory rules.

One way to think of the application of these rules, or rule schema (since they apply to an unbounded set of category pairs), is that the matching parts of the combining categories effectively cancel, leading to the rules being called \emph{cancellation laws} in some of the earlier work on categorial grammar. For example, when the lexical categories for \emph{Exchange} and \emph{Commission} in Figure~\ref{fig:application_deriv} are combined, the argument \cf{N} required by \emph{Exchange} in \cf{N/N} cancels with the lexical category \cf{N} for \emph{Commission}. We can also think of \cf{N/N} as a function that is applied to its argument \cf{N}. The \emph{forward} in \emph{forward application} refers to the fact that the argument is to the right. Backward ($<$) application---for when the argument is to the left---is used in the example when combining the subject \cf{NP} \emph{Investors} with the derived verb phrase \cf{S[dcl]\bs NP}.

\begin{figure}[t!]
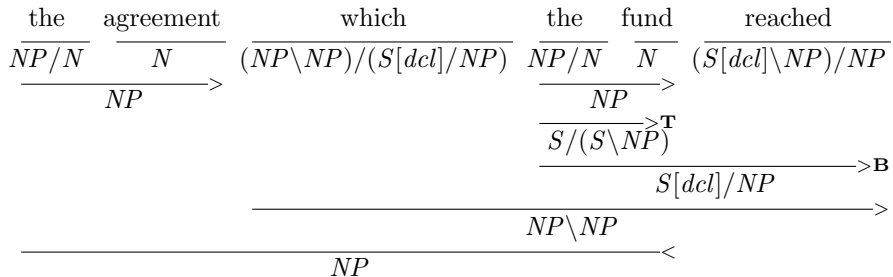


\deriv{6}{
\rm the & \rm agreement & \rm which & \rm the & \rm fund & \rm
reached\\
\uline{1}&\uline{1}&\uline{1}&\uline{1}&\uline{1}&\uline{1}\\
\cf{NP/N} & \cf{N} & \cf{(NP\bs NP)/(S[dcl]/NP)} & \cf{NP/N} & \cf{N} & \cf{(S[dcl]\bs NP)/NP}\\
\fapply{2} & & \fapply{2}\\
\mc{2}{\cf{NP}} && \mc{2}{\cf{NP}}\\
&&&\ftype{2}\\
&&&\mc{2}{\cf{S/(S\bs NP)}}\\
&&&\fcomp{3}\\
&&&\mc{3}{\cf{S[dcl]/NP}}\\
&&\fapply{4}\\
&&\mc{4}{\cf{NP\bs NP}}\\
\bapply{5}\\
\mc{5}{\cf{NP}}
}

\caption{Example derivation using type-raising and forward composition.}
\label{fig:fcomp}
\end{figure}

Figure~\ref{fig:fcomp} shows the derivation for a noun phrase containing a relative clause, where the object has been extracted out of its canonical position to the right of the transitive verb. The bracketing structure of the lexical category of the transitive verb (\cf{(S[dcl]\bs NP)/NP}) means that the verb is expecting to combine with its object to the right before its subject to the left. However, in this example the object has been moved away from the verb so that is not possible. The solution provided by CCG is to use two new combinatory rules. First, the unary rule of type-raising (${>}\mathbf{T}$) turns an atomic \cf{NP} category into a complex category \cf{S/(S\bs NP)}. A useful way to think about this new category is that it's a sentence missing a verb phrase (\cf{S\bs NP}) to the right, which is a natural way to conceive of a subject \cf{NP} as a function. Second, the rule of forward composition (${>}\mathbf{B}$) enables the combination of the type-raised noun phrase (\cf{S/(S\bs NP)}) and the transitive verb (\cf{S[dcl]\bs NP)/NP}), again with the idea that the verb-phrase categories ``in the middle" effectively cancel. This results in the slightly unusual constituent \cf{S[dcl]/NP}, which reflects the fact that the linguistic unit \emph{the fund reached} is a sentence missing an \cf{NP} to its right. Note that the lexical category for the relative pronoun in this example (\cf{(NP\bs NP)/(S[dcl]/NP)}) is expecting such a constituent to its right, so the relative pronoun can combine with the derived category using forward application.

There are additional combinatory rules in CCG which are designed to deal with other linguistic phenomena, including some rules in which the main slashes of the combining categories point in different directions -- the so-called ``non-harmonic" or crossing rules, such as backward crossed composition. These are all based on the operators of combinatory logic
\citep{curry:1958}; hence the term \emph{combinatory} in Combinatory Categorial Grammar. \cite{steedman:96}, \cite{steedman:2000} and \cite{baldridge:thesis02} contain many linguistic examples which motivate the particular set of rules in the theory.




There is much work on the formal properties of CCG, including the seminal papers of Vijay-Shanker, Weir and Joshi  in which it was proven that CCG is strictly more powerful than context-free grammars, but substantially less powerful than context-sensitive grammars  -- hence the term \emph{mildly} context-senstive \citep{weir:92}.  \cite{joshi:91} prove that CCG is weakly equivalent---i.e. generating the same string sets---to Tree Adjoining Grammar, Head Grammar, and Linear Indexed Grammar. This was a remarkable result given the apparent differences between these formalisms. Tree Adjoining Grammar \citep{joshi:87}, like CCG, has become a standard grammar formalism in Computational Linguistics and has formed the basis for much experimental work in developing parsers and NLP systems \citep{kasai-etal-2018-end}. \cite{kuhlmann-etal-2015-lexicalization} build on the earlier formal work and show that there are versions of CCG that are more powerful than CFGs, but strictly less powerful than TAG.
Despite the additional power of CCG (and TAG), there are still efficient parsing algorithms for CCG (and TAG) which are polynomial in the length of the input sentence \citep{vijay:93,kuhlmann-etal-2018-complexity}. 

The mildly context-sensitive nature of CGG is much trumpeted, and rightly so given that it enables analyses of the crossing dependencies in Dutch and Swiss German \citep{shieber:85}. However, it is perhaps worth pointing out that, for practical CCG parsing of English at least, the successful parsers have either used a CCG grammar which is context free by construction, being built entirely from rules instances observed in a finite CCG treebank \citep{hock:acl02,fowler-penn-2010-accurate}, or a grammar which is context free in practice by limiting the applicability of the combinatory rules to the rule instances in the treebank \citep{clark-curran-2007-wide}. Hence the parsing algorithms used by practical CCG parsers tend not to exploit the (somewhat complicated) structure-sharing schemes which define the more general polynomial-time parsing algorithms referenced above.

The remainder of this report starts out with the CCG supertagging task (Section~\ref{sec:supertagging}), showing the 20-year evolution of CCG supertagging from feature-based models in which the features are defined by hand, to neural models in which the features are induced automatically by a neural network. Section~\ref{sec:parsing} then demonstrates the gains that can be obtained by simply using a neural CCG supertagger as a front-end to an existing CCG parser, as well as additional improvements from using a neural classifier for the parsing model itself. Note that much of this report is a survey of existing work carried out by other researchers---or at least existing work replicated by the author---with the new material appearing in Section~\ref{sec:neural_parsing}, which reports new state-of-the-art accuracy figures for the CCGbank parsing task. It also acts as something of a survey of the 20-year wide-coverage CCG parsing project that began in Edinburgh.\footnote{https://groups.inf.ed.ac.uk/ccg/index.html} For a more detailed exposition of the linguistic theory of CCG, the reader is referred to \cite{steedman:96}, \cite{steedman:2000} and \cite{baldridge:thesis02}. For an introduction to wide-coverage CCG parsing, the reader is referred to \cite{clark-curran-2007-wide} and \cite{hockenmaier-steedman-2007-ccgbank}.

\section{CCG Supertagging}
\label{sec:supertagging}

CCG supertagging is the task of assigning a single lexical category (or ``supertag") to each word in an input sentence. The term \emph{supertag} originates from the seminal work of \cite{srinivas:99} for lexicalised tree-adjoining grammar (LTAG), and reflects the fact that CCG lexical categories (and elementary trees in LTAG) contain so much information. As an indication of how much information, note that the CCG grammar used in this report contains around 1,300 lexical categories, compared with the 50 or so part-of-speech tags in the original Penn Treebank \citep{marcus:93}.

\begin{figure}
\begin{center}
     \includegraphics[width=10cm]{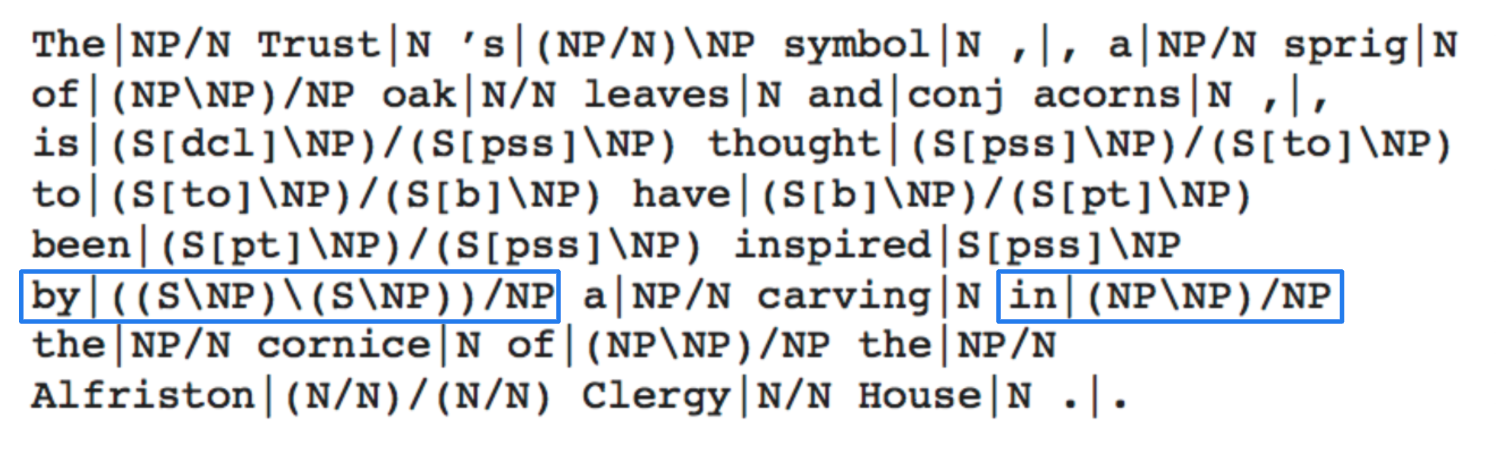}
    \caption{A sentence from Wikipedia with the correct lexical category assigned to each word.}
    \label{fig:wiki_stags}
\end{center}
\end{figure}

Figure~\ref{fig:wiki_stags} shows a sentence from Wikipedia with the correct lexical category assigned to each word \citep{clark_jhu:09}. The words highlighted in blue demonstrate why CCG supertagging is a difficult task. The lexical category assigned to \emph{by} takes two arguments: an \cf{NP} to the right and a verb phrase (\cf{S\bs NP}) to the left. The lexical category assigned to \emph{in} is similar, but takes an \cf{NP} to the left. In the Penn Treebank, both of these prepositions would be assigned the part-of-speech tag \cf{IN}. The point of the example is that, in order to assign the correct lexical category to these prepositions (or at least the second one), the supertagger has to decide whether the preposition attaches to a noun or a verb; i.e. it effectively has to resolve a prepositional phrase attachment ambiguity, which is one of the more difficult, classical parsing ambiguities \citep{collins-brooks-1995-prepositional}. This led \cite{srinivas:99} to describe supertagging as \emph{almost parsing}.

The data most used for training and testing CCG supertaggers is from CCGbank \citep{hockenmaier-steedman-2007-ccgbank}, which is a CCG version of the original Penn Treebank \citep{marcus:93}, a corpus of newswire sentences manually annotated with syntactic parse trees. A standard split is to take Sections 2-21 (39,604 sentences) as training data, Section 00 (1,913 sentences) as development data, and Section 23 (2,407 sentences) as test data. Extracting a grammar from Sections 2-21 results in 1,286 distinct lexical category types, with 439 of those types occurring only once in the training data.

The original CCG supertagger \citep{clark:tag02} was a maximum entropy (``maxent") tagger \citep{ratnaparkhi:96}, which was state-of-the-art for sequence labelling tasks at the time. The main difference with today's neural taggers is that the features were defined by hand, in terms of feature templates, based on linguistic intuition. For example, the NLP researcher may have decided that a good feature for deciding the correct tag for a word is the previous word in the sentence, which would then become a feature template which gets filled in for each particular word being tagged. Both types of tagger use iterative algorithms for training, typically maximising the likelihood of the (supervised) training data, with the neural models benefitting from specialised GPU hardware. Another difference is that the Transformer-based model described in Section~\ref{sec:super_neural} builds on a pre-trained model which has already been trained on large amounts of data to perform a fairly generic language modelling task. It turns out that this pre-training stage---which has become available since the maxent taggers because of developments in neural networks, hardware, and the availability of data---is crucial for the resulting performance of the supertagger, which is fine-tuned for the supertagging task.

The per-word accuracy for the maxent supertagger was around 92\% (see Table~\ref{tab:stagger_acc}), compared with over 96\% for the Penn Treebank pos-tagging task \citep{ratnaparkhi:96,curran-clark-2003-investigating}. An accuracy of 92\% may sound reasonable, given the difficulty of the task, but with an average sentence length in the treebank of around 20-25 words, this would result in approximately two errors every sentence. Given the amount of information in the lexical categories, it is crucial for the subsequent parsing stage that the supertagging is correct. Hence \cite{clark:coling04} developed a ``multitagging" approach in which the supertagger is allowed to dynamically assign more than one lexical category to a word, based on how certain the supertagger is of the category for that word. Allowing more than one lexical category increases the accuracy to almost 98\% with only a small increase in the average per-word lexical category ambiguity (see the end of Section~\ref{sec:super_neural} below).

Development of the maxent supertagger from 2007 mainly consisted of adapting it to other domains \citep{rimell-clark-2008-adapting} and using it to increase the speed of CCG parsers \citep{kummerfeld-etal-2010-faster}. The first paper to use neural methods for CCG supertagging was \cite{xu-etal-2015-ccg}, which applies a vanilla RNN to the sequence labelling task. This resulted in substantial accuracy improvements (Table~\ref{tab:stagger_acc}), and also produced a more robust supertagger that performed better on sentences from domains other than newswire (see the paper for details). Another improvement was that the maxent supertagger relied heavily on part-of-speech (POS) tags to define effective feature templates, and the accuracy degraded significantly when using automatically assigned, as opposed to gold-standard, POS tags (see the first two rows in Table~\ref{tab:stagger_acc}). The RNN supertagger was able to surpass the accuracy of the maxent supertagger relying on gold POS, but without using POS tags as input at all.

\begin{table}[t!]
\begin{center}
\begin{tabular}{llr}
\hline
$\textsc{Supertagger}$  &  \textsc{Model} & \textsc{Per-word acc}\\
\hline
\cite{clark-curran-2007-wide} & Maxent & 91.5 \\
\cite{clark-curran-2007-wide} (\emph{w/gold pos}) & Maxent & 92.6 \\
\cite{xu-etal-2015-ccg} & RNN &  93.1\\
\cite{lewis-etal-2016-lstm} & BiLSTM &  94.1\\
\cite{lewis-etal-2016-lstm} (+tri-training) & BiLSTM & 94.9 \\
\cite{tian-etal-2020-supertagging}  & Transformer & 96.2 \\
\hline
\end{tabular}
\end{center}
\caption{Supertagger accuracy evolution on Section 00 of CCGbank.}
\label{tab:stagger_acc}
\end{table}

\cite{lewis-etal-2016-lstm} improved on the performance of the vanilla RNN by using a bi-directional LSTM, and obtained additional gains by training on large amounts of automatically parsed data. More specifically, the tri-training method (row 5 in the table) uses the lexical category sequences from  parsed sentences as additional training data, but only those supertagged sentences on which two different CCG parsers agree. Finally, the last row of Table~\ref{tab:stagger_acc} shows the staggering improvements that can be had---over 50\% reduction in error rate compared to the original maxent model---when using a pre-trained neural language model that is fine-tuned for the supertagging task. This model is described in the next section.

Other recent neural approaches to CCG supertagging include \cite{clark-etal-2018-semi} and \cite{bhargava-penn-2020-supertagging}. \cite{bhargava-penn-2020-supertagging} is noteworthy because it uses neural sequence models to model the internal structure of lexical categories, which allows the supertagger to meaningfully assign non-zero probability mass to previously unseen lexical categories, as well as model rare categories, which is important given the long tail in the lexical category distribution.

\subsection{CCG Supertagging with a Transformer-based Model}
\label{sec:super_neural}

The field of NLP has experienced a period of rapid change in the last few years, due to the success of applying large-scale neural language models to a range of NLP tasks. The new paradigm relies on taking a pre-trained neural language model, which has been trained to carry out a fairly generic language modelling task such as predicting a missing word in a sentence, and fine-tuning it through additional supervised training for the task at hand \citep{NEURIPS2020_1457c0d6}. Despite the generic nature of the original language modelling task, the neural model is able to acquire large amounts of linguistic (and world) knowledge which can then be exploited for the downstream task. 

Another innovation which has been particularly influential is the development of the Transformer neural architecture, which consists of many self-attention layers, where every pair of words in a sentence is connected via a number of attention ``heads" \citep{vaswani2017attention,devlin-etal-2019-bert}. Each attention head calculates a similarity score between transformed representations of the respective word embeddings, where one word acts as a ``query" and the other as a ``key". These scores are then used by each word to derive a probabilistic mixture of all the other words in the sentence, where the mixture elements (``values") are again transformed representations of the word embeddings. This mixture acts as a powerful word-in-context representation, and stacking the attention layers a number of times, combined with some non-linear, fully-connected layers, results in a highly non-linear contextualised representation of each word in the sentence.

\begin{figure}
\begin{center}
     \includegraphics[width=12cm]{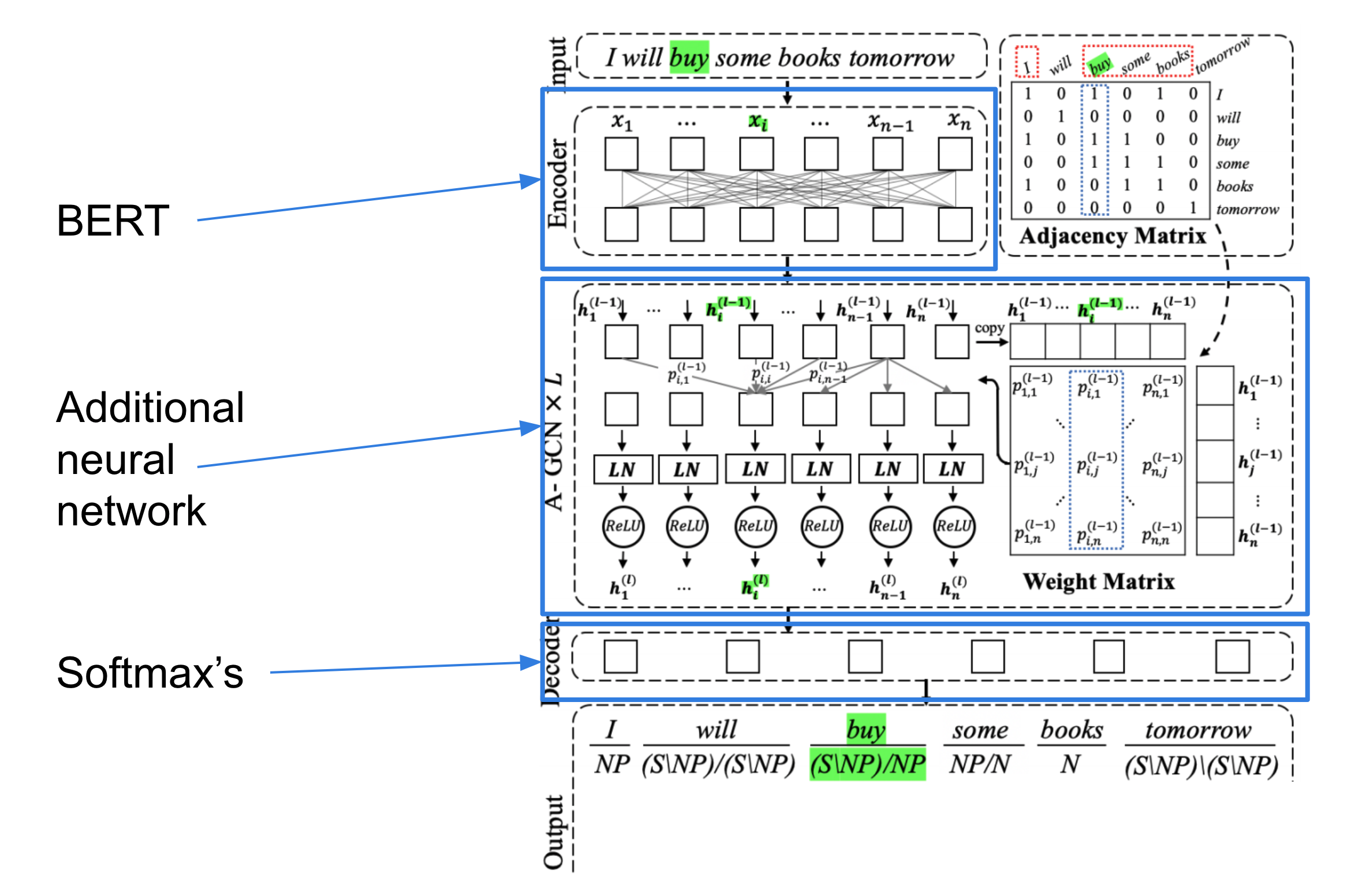}
    \caption{The neural architecture of the supertagger from \cite{tian-etal-2020-supertagging}.}
    \label{fig:tian_arch}
\end{center}
\end{figure}

\cite{tian-etal-2020-supertagging} apply this method to the CCG supertagging task, with great success. Figure~\ref{fig:tian_arch}, which is an embellished version of Figure 1 from the paper, shows the neural architecture. The first component is the BERT encoder \citep{devlin-etal-2019-bert}, which has already been pre-trained on large amounts of text using a masked language modelling objective. The output of BERT is a word embedding for each word in the input sentence. Then, an additional neural network takes the output of BERT, and also produces an embedding for each word. This additional network has not been pre-trained, and so requires supervised training data for its weights to be learned. In \cite{tian-etal-2020-supertagging} the additional network is a novel graph convolutional neural network (GCNN). Finally, there is an additional set of parameters which, for each output word embedding, define a softmax over the set of lexical categories. The output of the supertagger, for each input word, is the most probable lexical category according to the softmax distribution.

Training proceeds in a standard fashion, using the lexical category sequences from Sections 2-21 of CCGbank as supervised training data. The loss function is the cross-entropy loss, the minimisation of which is equivalent to putting as much probability mass as possible on each correct lexical category, relative to the incorrect lexical categories for each word (i.e. maximising the probability of the training data). A form of batch-based stochastic gradient descent is used to minimise the loss function, and dropout is used to prevent overfitting \citep{Goodfellow-et-al-2016}. All of these techniques have now become standard in neural NLP. The implementation uses the neural network library PyTorch.

One feature of the neural supertagger, compared with the maxent supertagger, is that it does not model the lexical category sequence at all. One of the challenges in developing taggers using sequence modelling methods, such as HMMs \citep{brants-2000-tnt}, CRFs \citep{laffertyCrf}, and maxent models, is that the number of possible tag sequences grows exponentially with sentence length, so modelling them explicitly requires either heuristic methods such as beam search, or dynamic programming techniques such as Viterbi. In contrast, the probabilistic decision made by the neural supertagger of what lexical category to assign to  each word is made independently of the decisions for the other words. The reason it performs so well is because of the highly contextualised nature of the output word embeddings, which already contain substantial amounts of information about the other words in the sentence.

There are a number of possibilities for the additional neural network in Figure~\ref{fig:tian_arch}. In fact, one possibility is not to add any additional layers at all, and simply fine-tune BERT, which works almost as well as the GCNN (Table 5 in \cite{tian-etal-2020-supertagging}). It is likely that adding in some additional attention layers, and training those with the supervised data, would work just as well.

In order to replicate the results in \cite{tian-etal-2020-supertagging}, and to use the supertagger as a front-end to an existing CCG parser, I downloaded and ran the code from the github repository\footnote{https://github.com/cuhksz-nlp/NeST-CCG}, retraining the supertagger on Sections 2-21 of CCGbank. One difference compared to \cite{tian-etal-2020-supertagging} is that I used the full lexical category set of 1,286 categories, rather than the 425 which result from applying a frequency cutoff. I also downloaded the 2019\_05\_30 BERT-Large Uncased model from the BERT repository\footnote{https://github.com/google-research/bert} to serve as the BERT encoder.

If the NeST-CCG supertagger is to act as an effective front-end to a CCG parser, it would be useful for it to sometimes output more than one category for a word. In fact, NeST-CCG already has a hyperparameter---\emph{clipping\_threshold}---which retains lexical categories based on their log-probabilities. Tuning this hyperparameter---let's call it $\gamma$---turns out to be highly effective for producing a multitagger. There is  a trade-off between using a low $\gamma$ value, which increases the chance of assigning the correct category, and a high $\gamma$ which reduces the average number of categories per word.  One of the motivations for supertagging is that, if the average number of categories per word can be kept low, then this greatly increases the efficiency of the parser \citep{clark-curran-2007-wide}.

The optimal $\gamma$ value depends on how ``sharp" the lexical category distributions are, and this is affected by the number of training epochs for the NeST-CCG supertagger. NeST-CCG also has an additional hyperparameter---call it $\alpha$---which sets a maximum number of lexical categories that can be assigned to a single word. I experimented with various combinations of $\gamma$, $\alpha$, and number of training epochs, and found a happy medium with $\gamma = 0.0005$, $\alpha = 10$, and 10 epochs, which resulted in a {\bf{multitagging accuracy of 99.3\%}} on the development data with {\bf{1.7 lexical categories per word}} on average.\footnote{In order for a set of lexical categories assigned to a word to be ``correct", the set needs to contain the one correct category. Hence multitagging ``accuracy" increases monotonically with smaller $\gamma$ values, as does the per-word ambiguity, reflecting the trade-off described above.} This compares very favourably with {\bf{97.6\%}} at {\bf{1.7 categories per word}} from \cite{clark-curran-2007-wide}. Hence the expectation is that this greatly improved multitagger will lead to improved parsing performance, which we turn to next.

\section{CCG Parsing}
\label{sec:parsing}

The job of a CCG parser is to take the output of a CCG supertagger as input, combine the categories together using the CCG combinatory rules, and return the best analysis as the output.  This process requires a parsing algorithm, which determines the order in which the categories are put together; a parsing model, which scores each possible analysis; and a search algorithm, which efficiently finds the highest-scoring analysis. In addition, there are various options for what kind of analysis the parser returns as output.

The most popular form of parsing algorithms for CCG have been bottom-up, in which the lexical categories are combined first, followed by combinations of categories with increasing spans, eventually resulting in a root category spanning the whole sentence \citep{steedman:2000}. The first parsing algorithms to be successfully applied to wide-coverage CCG parsing were chart-based \citep{hock:acl02,clark:acl04}, followed by shift-reduce parsers \citep{zhang-clark-2011-shift,xu-etal-2014-shift,ambati-etal-2016-shift}. In this report we will be using a chart-based parser (described in Section~\ref{sec:parsing_stagger_front-end}).

The original CCG parsing models were based on lexicalised PCFGs and used relative frequency counts to estimate the model parameters, with backing-off techniques to deal with data sparsity \citep{hock:acl02,collins:97}.\footnote{For a number of citation lists in this section, the first citations give the relevant CCG papers, followed by the (non-CCG) work on which they were based.} These were superseded by discriminative, feature-based models, essentially applying the maxent models that had worked so well for tagging to the parsing problem \citep{clark:acl04,riezler:acl02,miyao-tsujii-2008-feature}. Alternative estimation methods based on the structured perceptron framework---which provides a particularly simple estimation technique---were also applied successfully to CCG \citep{clark-curran-2007-perceptron,collins_roark:acl04}. The more recent neural parsing models that have been applied to CCG are mentioned in Section~\ref{sec:neural_parsing}.

In terms of search, the choice is between optimal dynamic programming, heuristic beam search, or optimal A* search \citep{lee-etal-2016-global}. The early CCG parsing work focused on dynamic programming algorithms \citep{hock:acl02,clark:acl04}, whereas the shift-reduce CCG parsers tended to use beam search, performing surprisingly well even with relatively small beam widths \citep{zhang-clark-2011-shift}. The CCG parser described below also uses beam search, but applied to a chart.

Finally, the main output formats have been either the CCG derivation itself, a dependency graph where the dependency types are defined in terms of the CCG lexical categories \citep{clark:acl02,hock:acl02}, or a dependency graph using a fairly formalism-independent representation \citep{clark-curran-2007-wide}. It has been argued that dependency types are especially useful for parser evaluation \citep{carroll:98}, in particular for CCG \citep{clark:lrec02}, as well as for downstream NLP tasks and applications. In this report the parser output will be CCG dependencies (used for evaluation), with a novel application of the derivations suggested in the Conclusion. 
There is also a large body of work on interpreting CCG derivations to produce semantic representations as logical forms, in particular for Discourse Representation Theory \citep{bos:coling04,bos:2017,bos:21}, Abstract Meaning Representation \citep{artzi-etal-2015-broad}, as well as for more general semantic parsing tasks \citep{zettlemoyer:05,artzi-etal-2014-learning}.

\subsection{CCG Parsing with a Neural Supertagger Front End}
\label{sec:parsing_stagger_front-end}

This section describes the accuracy gains that can be obtained by simply using the neural supertagger described in Section~\ref{sec:super_neural} as a front-end to an existing CCG parser. These experiments further demonstrate the importance of supertagging for CCG \citep{clark:coling04}, and further realise the original vision of \cite{srinivas:99} for supertagging as almost parsing.

The CCG parser that we will use is the Java C\&C parser described in \cite{java_candc}. This is essentially a Java reimplementation of the C\&C parser, which was highly optimised C++ code designed for efficiency \citep{clark-curran-2007-wide,curran-etal-2007-linguistically}; one of the aims of the reimplementation was to make the code more readable and easier to modify. There were also some improvements made to the grammar, by extending the lexical category set that can be handled by the parser from the 425 lexical categories in C\&C to the full set of 1,286 derived from Sections 2-21 of CCGbank.\footnote{Extending the grammar in this way required an extension of the so-called \emph{markedup} file, which encodes how CCG dependencies are generated from the lexical categories.} Also, Java C\&C uses a new chart-based beam-search decoder, which removes any restrictions imposed by dynamic programming algorithms on the locality of the model features (at the cost of optimality), together with the max-violation framework of \cite{huang-etal-2012-structured} for training the linear model. And finally, Java C\&C has a new ``skimmer" mode which enables the parser to return a dependency analysis even when there is no full spanning derivation of the input sentence in the chart.

\begin{table}[t!]
\begin{center}
\begin{tabular}{llrrrrr}
\hline
$\textsc{Parser}$  &  \textsc{S-tagger} & \textsc{P} & \textsc{R} & \textsc{F} & \textsc{Cat} & \textsc{Cov.}\\
\hline
C\&C & Maxent & --\hspace*{2.5mm} & --\hspace*{2.5mm} & 85.3 & --\hspace*{2.5mm} & 99.1 \\
C\&C (\emph{w/gold pos}) & Maxent & 88.1 & 86.4 & 87.2 & 94.2 & 99.1 \\
Java C\&C (\emph{w/gold pos}) & Maxent &  88.0 & 87.3 & 87.7 & 94.3 & 100.0\\
Java C\&C & Transformer & 91.9 & 91.5 & 91.7 & 96.3 & 100.0\\
\hline
\end{tabular}
\end{center}
\caption{Parser accuracy of (Java) C\&C on Sec. 00 with different supertaggers.}
\label{tab:candc_acc}
\end{table}

Table~\ref{tab:candc_acc} shows the accuracy on the development data of the original C\&C parser and the Java C\&C parser, the latter with both the  maxent supertagger and the neural supertagger. The parser accuracy scores are the standard measures of labelled precision and recall over CCG dependencies, F-score (harmonic mean of precision and recall), and lexical category accuracy. The coverage figure is the percentage of sentences in Section 00 used for the evaluation. The figures for the C\&C parser are taken from \cite{clark-curran-2007-wide}, and those for Java C\&C with the maxent supertagger from \cite{java_candc}.\footnote{The new figures for Java C\&C were obtained using the \emph{evaluate\_new} script from the Java C\&C codebase.}

Moving from row 1 to 2 in the table, we again see the reliance on POS tags as features, for both the maxent supertagger and the original C\&C parsing model. Using automatically-assigned POS tags reduces the overall F-score by almost 2\%. Row 3 shows the improvements that are obtained with Java C\&C, both in terms of accuracy, but also in terms of coverage, since the parser now returns an analysis for all the sentences in the development data (using the ``skimmer" mode mentioned above). Finally, row 4 shows the staggering improvements that can be obtained by simply replacing the maxent supertagger with the neural supertagger. The coverage is again 100\% through use of the skimmer. These figures were obtained using a beam size of 32 in the parser, and the default parser model that ships with the Java C\&C codebase (and using automatically-assigned POS tags as features in the model).\footnote{This model has 8,137,733 features, although only 2.3M of these have non-zero weights after training. The logit scores from the supertagger were also used as (real-valued) features in the parser model, with a weighting of 10.0 set manually.} The settings on the supertagger were $\gamma = 0.0005$, $\alpha = 10$, and the supertagger model had been trained for 10 epochs, as described in Section~\ref{sec:super_neural}.

\subsection{CCG Parsing with a Transformer-based Model}
\label{sec:neural_parsing}

The parser model used in Section~\ref{sec:parsing_stagger_front-end} is a linear model based on discrete feature templates defined by hand, which raises the obvious question of whether additional gains can be obtained by using a neural model in the parser, as well as the supertagger. Some neural models for CCG parsing have already appeared in the literature \citep{xu-etal-2016-expected,xu-2016-lstm,lee-etal-2016-global,ambati-etal-2016-shift,stanojevic-steedman-2019-ccg}, but none of these use the pre-trained plus fine-tuning Transformer-based paradigm that was so successful for the supertagger. 

\begin{figure}
\begin{center}
     \includegraphics[width=12cm]{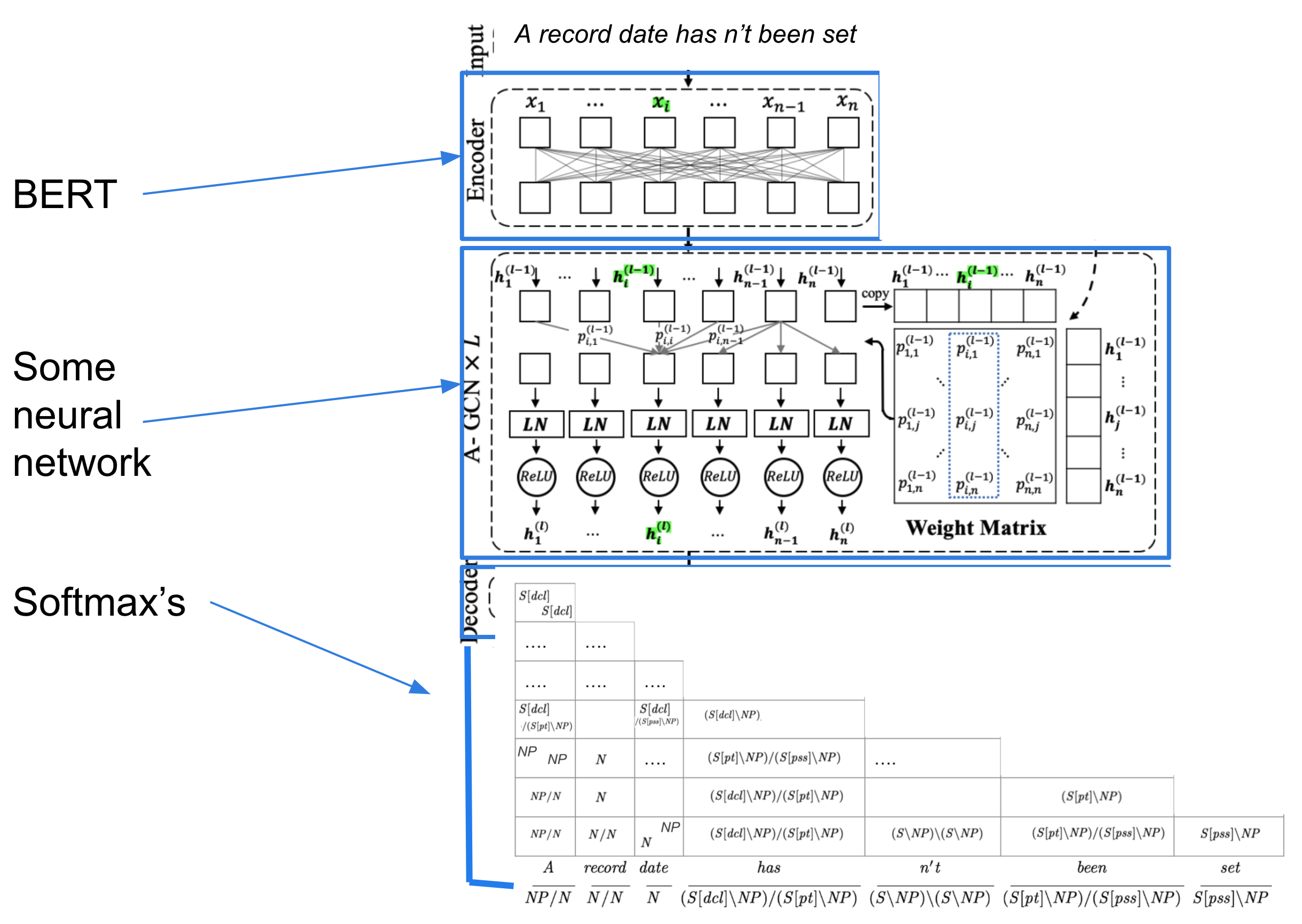}
    \caption{The architecture from \cite{tian-etal-2020-supertagging} adapted for chart parsing.}
    \label{fig:neural_chart}
\end{center}
\end{figure}

The difficulty with applying a similar technique to a parse chart is that scores need to be defined for each cell in the chart (corresponding to each span in the sentence), whereas the neural architecture only produces output embeddings for each word. Figure~\ref{fig:neural_chart} gives the general idea (adapted from Figure~\ref{fig:tian_arch} for the supertagger). How can the output embeddings for each word be used to produce an embedding for each span? In fact, this problem has already been solved for dependency parsing by \cite{wang-chang-2016-graph}, and successfully applied to the Penn Treebank constituency parsing task by \cite{stern-etal-2017-minimal} and \cite{kitaev-klein-2018-constituency}, in the latter case using a fine-tuned BERT encoder. The idea is to produce an embedding for a span by \emph{subtracting} the embedding for the word at the beginning of the span 
from the embedding of the word at the end. This is highly intuitive for the left-to-right LSTM used by \cite{wang-chang-2016-graph} (and in the bidirectional setting, the subtraction would be performed in the opposite order for the right-to-left LSTM); however, it is less intuitive for the Transformer, which has no encoding direction. Even so, \cite{stern-etal-2017-minimal} were able to successfully adapt this method for the Transformer.

Now that we have a method of producing an embedding for each cell, an extra set of parameters can be used to define a softmax over all possible constituent labels in the chart. This softmax can then be applied to each cell, producing a list of label-score pairs, with the score for a derivation defined as the sum of the scores (log-probabilities) for each node in the derivation. \cite{kitaev-klein-2018-constituency} used these scores to define a Viterbi-style dynamic programming algorithm for finding the highest-scoring derivation. Note that, as for the supertagger case, the scores for each cell are calculated independently of all the other cells (i.e. there is no explicit statistical modelling of the tree structure). The reason the model is so effective, despite this very strong independence assumption, is because of the highly contextualised nature of the word embeddings used to derive the cell embeddings. However, for the CCG case (and unlike the Penn Treebank parsing problem), there is a twist: any categories that are selected by a parsing algorithm to form a derivation must conform to the combinatory rules of CCG, and the independence assumption means that the highest-scoring tree may not satisfy this constraint.

\cite{kato:emnlp21} solve this problem by defining a new representation for a CCG constituent so that the items being scored conform to the CCG grammar.\footnote{A useful side-effect of the new representation is the possibility of modelling unseen categories, a feature shared with \cite{bhargava-penn-2020-supertagging}.} Here I adopt a much simpler method: only apply the score function to derivations that are produced by the grammar. That way, even if a category receives a high score for a span, it will only appear in the output if it forms part of a high-scoring, legal CCG derivation. Moreover, adapting the chart-based beam-search algorithm to use the new scores is particularly straightforward: simply replace the scores for a sub-derivation previously calculated according to the feature-based model with the scores from the neural model (where the score for a sub-derivation is the sum of the scores of the category labels on each node).

I downloaded the codebase for \cite{kitaev-klein-2018-constituency} and \cite{kitaev-etal-2019-multilingual} from the github repository,\footnote{https://github.com/nikitakit/self-attentive-parser} and trained a CCG parsing model by running the training code on the derivations in Sections 2-21 of CCGbank. Following \cite{kitaev-klein-2018-constituency}, unary chains in the CCG derivations---which arise as a result of type-raising and the unary type-changing rules in CCGbank \citep{hockenmaier-steedman-2007-ccgbank}---were represented as a single label.
As explained above, it is not possible to simply run the corresponding parser code at test time, since this may not produce legal CCG derivations, which makes it difficult to produce dependency structures for evaluation. Hence I modified the test-time parsing code to output the scores for each possible category in each cell,\footnote{The category set derived from Sections 2-21 contains 964 categories (including the unary chains, and excluding lexical categories).} and modified the Java C\&C code so that it is able to accept these charts as input, which now provide the scores for the beam-search decoder. As before, the parser model also uses the log-probabilities from the supertagger as additional scores, but this time weighted equally with the span-based scores (so the score for a derivation is the sum of the scores on each node, including the lexical categories at the leaves).

\begin{table}[t!]
\begin{center}
\begin{tabular}{llrrrrr}
\hline
$\textsc{Parser}$  &  \textsc{S-tagger/} & \textsc{P} & \textsc{R} & \textsc{F} & \textsc{Cat} & \textsc{Cov.}\\
& \textsc{Parser Model} \\
\hline
C\&C & Maxent/Maxent & 86.2 & 84.7 & 85.5 & 93.0 & 99.6 \\
C\&C (\emph{w/gold pos}) & Maxent/Maxent & 88.3 & 87.0 & 87.6 & 94.3 & 99.6 \\
Java C\&C & T-former/Linear & 92.0 & 91.8 & 91.9 & 96.4 & 100.0 \\
Java C\&C & T-former/T-former &  92.8 & 93.0 & 92.9 & 96.5 & 100.0 \\
\hline
\end{tabular}
\end{center}
\caption{Parser accuracy of Java C\&C on Sec. 23 with neural models for both supertagger and parser, compared to other supertagger-parser combinations.}
\label{tab:parser_test_results}
\end{table}

\begin{table}[t!]
\begin{center}
\begin{tabular}{llrr}
\hline
$\textsc{Paper}$  &  \textsc{S-tagger/Parser}  & \textsc{F} & \textsc{Cat} \\
\hline
\cite{xu-2016-lstm} & LSTM/Shift-Reduce(LSTM) & 87.8 & 94.6\\
\cite{lewis-etal-2016-lstm} & LSTM/A* & 88.1 & --\hspace*{2.5mm}\\
\cite{vaswani-etal-2016-supertagging} & LSTM/Java C\&C & 88.3 & --\hspace*{2.5mm} \\
\cite{lee-etal-2016-global} & LSTM/A*(Tree-LSTM) & 88.7 & --\hspace*{2.5mm} \\
\cite{yoshikawa-etal-2017-ccg} & LSTM/A*(LSTM) & 88.8 & --\hspace*{2.5mm} \\
\cite{stanojevic-steedman-2020-max} & LSTM/Shift-Reduce(LSTM) & 90.6 & 95.6 \\
\cite{tian-etal-2020-supertagging} & T-former/EasyCCG &  90.7 & 96.4 \\
This report & T-former/Java C\&C &  92.9 & 96.5 \\
& \hspace*{1.5cm}(w/T-former)\\
\hline
\end{tabular}
\end{center}
\caption{Parser accuracy of Java C\&C on Sec. 23 with neural models for both supertagger and parser, compared to some recent CCG parsers in the literature.}
\label{tab:parser_league_table}
\end{table}

Table~\ref{tab:parser_test_results} shows the performance of the new parser on the Section 23 test set, compared with the other C\&C parsers, which use various combinations of supertagger and parser models. Replacing the linear parsing model based on hand-defined feature templates with the span-based neural model resulted in an absolute improvement of 1.0 F-score, and the improvement over the original C\&C model represents a reduction in error rate of more than 50\%. Moreover, the percentage of sentences that are completely correct is now 54.1 (not shown in the table), compared with 32.9\% for the original C\&C parser.

Table~\ref{tab:parser_league_table} shows the performance of the Java C\&C parser with neural models compared with recent CCG parsing results from the literature, where all the accuracy figures are taken from the respective papers. The final F-score of 92.9 on the test set represents a new state-of-the-art for the CCGbank parsing task.

Table~\ref{table:reln_res} shows the accuracy of the parser by dependency relation, with the table copied from \cite{clark-curran-2007-wide}; the new figures for this report are in the final F column, and the \emph{\#deps} column gives the number of dependencies of the corresponding type in Section 00 (recalculated to reflect the 100\% coverage with Java C\&C). Substantial improvements over the original C\&C parser can be seen across the board. Of particular note are the improvements for \emph{np modifying prep} and \emph{vp modifying prep}, which are dependencies potentially resulting from PP-attachment ambiguities, and the 100\% accuracy scores for \emph{object rel pronoun}, which are long-range dependencies resulting from extraction out of a relative clause and particularly difficult for parsers to recover correctly \citep{rimell-etal-2009-unbounded}.

\begin{table}
\setlength{\tabcolsep}{1.2mm}
\small
\begin{tabular}{lclrrr}\hline 
Lexical category & Arg Slot & & $\# deps$ & F(C\&C) & F \\
\hline
$\cf{N}/\cf{N}\head{1}$ & $1$ & \reln{nominal modifier} &
  $7\,443$ & 95.5 & 98.1\\
$\cf{NP}/\cf{N}\head{1}$ & $1$ & \reln{determiner} & $4\,153$ & 96.3 & 98.8\\
$(\cf{NP}\bs\cf{NP}\head{1})/\cf{NP}\head{2}$ & $2$
& \reln{np modifying prep} & 2\,390 &  85.4 & 94.3\\
$(\cf{NP}\bs\cf{NP}\head{1})/\cf{NP}\head{2}$ & $1$
& \reln{np modifying prep}  & 2\,193 & 83.6 & 93.9\\
$((\cf{S}\bs\cf{NP})\bs(\cf{S}\head{1}\bs\cf{NP}))/\cf{NP}\head{2}$ 
& $2$ & \reln{vp modifying prep}  & 1\,177 &  72.6 & 85.6\\
$((\cf{S}\bs\cf{NP})\bs(\cf{S}\head{1}\bs\cf{NP}))/\cf{NP}\head{2}$
& $1$ & \reln{vp modifying prep} &  1\,068 & 71.4 & 84.2\\
$(\cf{S}\feat{dcl}\bs\cf{NP}\head{1})/\cf{NP}\head{2}$ & $1$ &
\reln{transitive verb} & 877 &  83.5 & 95.5\\
$(\cf{S}\feat{dcl}\bs\cf{NP}\head{1})/\cf{NP}\head{2}$ & $2$ &
\reln{transitive verb} & 923 & 83.9 & 96.0\\
$(\cf{S}\bs\cf{NP})\bs(\cf{S}\head{1}\bs\cf{NP})$
& $1$ & \reln{adverbial modifier} &  750 & 86.8 & 93.0\\
$\cf{PP}/\cf{NP}\head{1}$
& $1$ & \reln{prep complement} & 885 & 72.5 & 86.4\\
$(\cf{S}\feat{b}\bs\cf{NP}\head{1})/\cf{NP}\head{2}$
& $2$ & \reln{inf transitive verb}  & 648 & 85.5 & 94.8\\
$(\cf{S}\feat{dcl}\bs\cf{NP}\head{1})/(\cf{S}\feat{b}\head{2}\bs\cf{NP})$
& $2$ & \reln{auxiliary} & 480 & 97.8 & 99.2\\
$(\cf{S}\feat{dcl}\bs\cf{NP}\head{1})/(\cf{S}\feat{b}\head{2}\bs\cf{NP})$
& $1$ & \reln{auxiliary} & 488 &  93.6 & 98.0\\
$(\cf{S}\feat{b}\bs\cf{NP}\head{1})/\cf{NP}\head{2}$
& $1$ & \reln{inf transitive verb} & 529 & 75.8 & 90.5\\
$(\cf{NP}/\cf{N}\head{1})\bs\cf{NP}\head{2}$ & $1$ & \reln{s 
  genitive}  & 385 &  96.1 & 97.9\\
$(\cf{NP}/\cf{N}\head{1})\bs\cf{NP}\head{2}$ & $2$ & \reln{s 
  genitive} & 374 & 98.0 & 99.2\\
$(\cf{S}\feat{dcl}\bs\cf{NP}\head{1})/\cf{S}\feat{dcl}\head{2}$
& $1$ & \reln{sentential comp verb} & 387 & 92.9 & 98.3\\
$(\cf{NP}\bs\cf{NP}\head{1})/(\cf{S}\feat{dcl}\head{2}\bs\cf{NP})$ & $1$
& \reln{subject rel pronoun} & 277 & 83.5 & 93.8\\
$(\cf{NP}\bs\cf{NP}\head{1})/(\cf{S}\feat{dcl}\head{2}\bs\cf{NP})$ & $2$
& \reln{subject rel pronoun} & 279 &  97.3 & 99.1 \\
$(\cf{NP}\bs\cf{NP}\head{1})/(\cf{S}\feat{dcl}\head{2}/\cf{NP})$
& $1$ & \reln{object rel pronoun} & 26 & 75.0 & 100.0\\
$(\cf{NP}\bs\cf{NP}\head{1})/(\cf{S}\feat{dcl}\head{2}/\cf{NP})$
& 2 &\reln{object rel pronoun} & 23 & 84.4 & 100.0\\
$\cf{NP}/(\cf{S}\feat{dcl}\head{1}/\cf{NP})$
& $1$ & \reln{headless obj rel pron} &17 & 100.0 & 100.0\\
\hline
\end{tabular}
\caption{Parser accuracy on Section 00 by dependency relation.}
\label{table:reln_res}
\end{table}

Finally, Table~\ref{tab:parser_speeds} shows how the parser accuracy and parser speed varies with beam width. Accuracies are reported for Section 00, and the approximate speeds are calculated using the first 1,000 sentences of Section 00, running on a linux server.\footnote{Model Intel Xeon W-2265 CPU @ 3.50GHz.} The \%-skimmed column gives the percentage of sentences that are analysed using the skimmer mode of the parser.\footnote{This percentage goes up and down with beam width because there are two parameters which affect it: the beam width itself, since a smaller beam reduces the chances of the chart containing a spanning derivation; and the maximum chart size allowed by the parser, which will get hit more often---hence triggering the skimmer---with a larger beam.} The parser is remarkably resilient to a decreasing beam width, with accuracy and coverage only being substantially affected with beam widths as low as 4. The ability of neural models to effectively utilise a large context when making decisions---thereby reducing the need for search---has been seen already for parsing, especially for shift-reduce parsing where even fully-greedy parsers can be highly accurate \citep{chen-manning-2014-fast,xu-etal-2016-expected}. The speeds themselves are eclipsed by the parser of \cite{lewis-etal-2016-lstm}, which can parse over 2,000 sentences a second by exploiting the efficiency of the A* parsing algorithm and running the supertagger in parallel on a GPU. However, the speeds reported in Table~\ref{tab:parser_speeds} are reasonable considering that the implementation is in Java and has not been optimised, and the parser was run on the CPU; these speeds of around 100 sentences per second are still high enough to be used for large-scale textual data analysis.

\begin{table}[t!]
\begin{center}
\begin{tabular}{rrrr}
\hline
$\textsc{Beam-width}$  & \textsc{F (Sec. 00)} & \textsc{Sents/sec} & \textsc{\%-skimmed}\\
\hline
64 & 93.0 & 8 & 0.9\\
32 & 93.0 & 23 &  0.7 \\
16 & 93.0 & 51 & 0.5 \\
8 &  92.9 & 105 &  0.4 \\
4 & 91.7 & 167 & 2.2\\
\hline
\end{tabular}
\end{center}
\caption{The speed/accuracy trade-off with different beam sizes.}
\label{tab:parser_speeds}
\end{table}

\section{Conclusion}

This report has shown how a combination of Transformer-based neural models and a symbolic CCG grammar can lead to substantial gains over existing approaches to CCG parsing. The staggering improvements compared to the original C\&C parser \citep{clark:acl04}, with overall dependency accuracies of almost 93\% and complete sentence accuracies of  54\%, are representative of how NLP has evolved over this time. In terms of how useful such parsers are for downstream tasks and applications, the success of large-scale language models \citep{NEURIPS2020_1457c0d6} has, for now at least, largely replaced the old NLP pipeline in which parsers played a prominent role. Whether this trend will continue, and what this trend means for the nature of linguistic representation and learning, remain open questions \citep{lappin:21}. 
However, even if this trend does continue, there may still be unforeseen roles for structural analysis in NLP; for example at Cambridge Quantum we are investigating how parsers can be used to build quantum circuits, exploiting the similarities between the compositional structures in categorial grammar and those in vector spaces, the latter of which lie at the heart of quantum computation \citep{coecke:2010,yeung:21,lorenz:21}.

\section*{Acknowledgments}

Thanks to the developers of the \cite{tian-etal-2020-supertagging} and \cite{kitaev-etal-2019-multilingual} codebases for making their code freely available and so easy to use. Thanks to Mark Steedman and Julia Hockenmaier for starting the Edinburgh CCG parsing project all those years ago, and to James Curran for the great memories developing the C\&C parser. Thanks to the collaborators and students down the years who have worked on the parser, including Laura Rimell, Yue Zhang, Wenduan Xu, Luana Bulat, and Darren Foong. And finally, thanks to Shalom Lappin; Ilyas Khan; Konstantinos Meichanetzidis, Dimitri Kartsaklis and the rest of the Compositional Intelligence team at Cambridge Quantum for providing useful feedback on this latest work.

\bibliography{clark}

\begin{thebibliography}{81}
\expandafter\ifx\csname natexlab\endcsname\relax\def\natexlab#1{#1}\fi

\bibitem[Ajdukiewicz(1935)]{ajdukiewicz:35}
Ajdukiewicz, Kazimierz. 1935.
\newblock Die syntaktische konnexit\"{a}t.
\newblock {\em Studia Philosophica\/} 1:1--27.

\bibitem[Ambati et~al.(2016)Ambati, Deoskar, and
  Steedman]{ambati-etal-2016-shift}
Ambati, Bharat~Ram, Tejaswini Deoskar, and Mark Steedman. 2016.
\newblock Shift-reduce {CCG} parsing using neural network models.
\newblock In {\em Proceedings of the 2016 Conference of the North {A}merican
  Chapter of the Association for Computational Linguistics: Human Language
  Technologies\/}, pages 447--453. San Diego, California: Association for
  Computational Linguistics.

\bibitem[Artzi et~al.(2014)Artzi, Das, and Petrov]{artzi-etal-2014-learning}
Artzi, Yoav, Dipanjan Das, and Slav Petrov. 2014.
\newblock Learning compact lexicons for {CCG} semantic parsing.
\newblock In {\em Proceedings of the 2014 Conference on Empirical Methods in
  Natural Language Processing ({EMNLP})\/}, pages 1273--1283. Doha, Qatar:
  Association for Computational Linguistics.

\bibitem[Artzi et~al.(2015)Artzi, Lee, and Zettlemoyer]{artzi-etal-2015-broad}
Artzi, Yoav, Kenton Lee, and Luke Zettlemoyer. 2015.
\newblock Broad-coverage {CCG} semantic parsing with {AMR}.
\newblock In {\em Proceedings of the 2015 Conference on Empirical Methods in
  Natural Language Processing\/}, pages 1699--1710. Lisbon, Portugal:
  Association for Computational Linguistics.

\bibitem[Baldridge(2002)]{baldridge:thesis02}
Baldridge, Jason. 2002.
\newblock {\em Lexically Specified Derivational Control in Combinatory
  Categorial Grammar\/}.
\newblock Ph.D. thesis, Edinburgh University.

\bibitem[Bangalore and Joshi(1999)]{srinivas:99}
Bangalore, Srinivas and Aravind Joshi. 1999.
\newblock Supertagging: An approach to almost parsing.
\newblock {\em Computational Linguistics\/} 25(2):237--265.

\bibitem[Bar-Hillel(1953)]{bar-hillel:53}
Bar-Hillel, Yehoshua. 1953.
\newblock A quasi-arithmetical notation for syntactic description.
\newblock {\em Language\/} 29:47--58.

\bibitem[Bhargava and Penn(2020)]{bhargava-penn-2020-supertagging}
Bhargava, Aditya and Gerald Penn. 2020.
\newblock Supertagging with {CCG} primitives.
\newblock In {\em Proceedings of the 5th Workshop on Representation Learning
  for NLP\/}, pages 194--204. Online: Association for Computational
  Linguistics.

\bibitem[Bos et~al.(2017)Bos, Basile, Evang, Venhuizen, and Bjerva]{bos:2017}
Bos, Johan, Valerio Basile, Kilian Evang, NJ~Venhuizen, and Johannes Bjerva.
  2017.
\newblock The groningen meaning bank.
\newblock In N.~Ide and J.~Pustejovsky, eds., {\em Handbook of Linguistic
  Annotation\/}, pages 463--496. Springer.

\bibitem[Bos et~al.(2004)Bos, Clark, Steedman, Curran, and
  Hockenmaier]{bos:coling04}
Bos, Johan, Stephen Clark, Mark Steedman, James~R. Curran, and Julia
  Hockenmaier. 2004.
\newblock Wide-coverage semantic representations from a {C}{C}{G} parser.
\newblock In {\em Proceedings of COLING-04\/}, pages 1240--1246. Geneva,
  Switzerland.

\bibitem[Brants(2000)]{brants-2000-tnt}
Brants, Thorsten. 2000.
\newblock {T}n{T} {--} a statistical part-of-speech tagger.
\newblock In {\em Sixth Applied Natural Language Processing Conference\/},
  pages 224--231. Seattle, Washington, USA: Association for Computational
  Linguistics.

\bibitem[Brown et~al.(2020)Brown, Mann, Ryder, Subbiah, Kaplan, Dhariwal,
  Neelakantan, Shyam, Sastry, Askell, Agarwal, Herbert-Voss, Krueger, Henighan,
  Child, Ramesh, Ziegler, Wu, Winter, Hesse, Chen, Sigler, Litwin, Gray, Chess,
  Clark, Berner, McCandlish, Radford, Sutskever, and
  Amodei]{NEURIPS2020_1457c0d6}
Brown, Tom, Benjamin Mann, Nick Ryder, Melanie Subbiah, Jared~D Kaplan,
  Prafulla Dhariwal, Arvind Neelakantan, Pranav Shyam, Girish Sastry, Amanda
  Askell, Sandhini Agarwal, Ariel Herbert-Voss, Gretchen Krueger, Tom Henighan,
  Rewon Child, Aditya Ramesh, Daniel Ziegler, Jeffrey Wu, Clemens Winter, Chris
  Hesse, Mark Chen, Eric Sigler, Mateusz Litwin, Scott Gray, Benjamin Chess,
  Jack Clark, Christopher Berner, Sam McCandlish, Alec Radford, Ilya Sutskever,
  and Dario Amodei. 2020.
\newblock Language models are few-shot learners.
\newblock In H.~Larochelle, M.~Ranzato, R.~Hadsell, M.~F. Balcan, and H.~Lin,
  eds., {\em Advances in Neural Information Processing Systems\/}, vol.~33,
  pages 1877--1901. Curran Associates, Inc.

\bibitem[Carroll et~al.(1998)Carroll, Briscoe, and Sanfilippo]{carroll:98}
Carroll, John, Ted Briscoe, and Antonio Sanfilippo. 1998.
\newblock Parser evaluation: a survey and a new proposal.
\newblock In {\em Proceedings of the 1st LREC Conference\/}, pages 447--454.
  Granada, Spain.

\bibitem[Chen and Manning(2014)]{chen-manning-2014-fast}
Chen, Danqi and Christopher Manning. 2014.
\newblock A fast and accurate dependency parser using neural networks.
\newblock In {\em Proceedings of the 2014 Conference on Empirical Methods in
  Natural Language Processing ({EMNLP})\/}, pages 740--750. Doha, Qatar:
  Association for Computational Linguistics.

\bibitem[Chomsky(1965)]{chomsky:65}
Chomsky, Noam. 1965.
\newblock {\em Aspects of the Theory of Syntax\/}.
\newblock MIT Press.

\bibitem[Clark et~al.(2018)Clark, Luong, Manning, and Le]{clark-etal-2018-semi}
Clark, Kevin, Minh-Thang Luong, Christopher~D. Manning, and Quoc Le. 2018.
\newblock Semi-supervised sequence modeling with cross-view training.
\newblock In {\em Proceedings of the 2018 Conference on Empirical Methods in
  Natural Language Processing\/}, pages 1914--1925. Brussels, Belgium:
  Association for Computational Linguistics.

\bibitem[Clark(2002)]{clark:tag02}
Clark, Stephen. 2002.
\newblock A supertagger for {C}ombinatory {C}ategorial {G}rammar.
\newblock In {\em Proceedings of the TAG+ Workshop\/}, pages 19--24. Venice,
  Italy.

\bibitem[Clark et~al.(2009)Clark, Copestake, Curran, Zhang, Herbelot, Haggerty,
  Ahn, Wyk, Roesner, Kummerfeld, and Dawborn]{clark_jhu:09}
Clark, Stephen, Ann Copestake, James~R. Curran, Yue Zhang, Aurelie Herbelot,
  James Haggerty, Byung-Gyu Ahn, Curt~Van Wyk, Jessika Roesner, Jonathan
  Kummerfeld, and Tim Dawborn. 2009.
\newblock Large-scale syntactic processing: Parsing the web.
\newblock Tech. rep., Center for Language and Speech Processing, Johns Hopkins
  University, Baltimore, MD.

\bibitem[Clark and Curran(2007{\natexlab{a}})]{clark-curran-2007-perceptron}
Clark, Stephen and James Curran. 2007{\natexlab{a}}.
\newblock Perceptron training for a wide-coverage lexicalized-grammar parser.
\newblock In {\em {ACL} 2007 Workshop on Deep Linguistic Processing\/}, pages
  9--16. Prague, Czech Republic: Association for Computational Linguistics.

\bibitem[Clark and Curran(2004{\natexlab{a}})]{clark:coling04}
Clark, Stephen and James~R. Curran. 2004{\natexlab{a}}.
\newblock The importance of supertagging for wide-coverage {C}{C}{G} parsing.
\newblock In {\em Proceedings of COLING-04\/}, pages 282--288. Geneva,
  Switzerland.

\bibitem[Clark and Curran(2004{\natexlab{b}})]{clark:acl04}
Clark, Stephen and James~R. Curran. 2004{\natexlab{b}}.
\newblock Parsing the {W}{S}{J} using {C}{C}{G} and log-linear models.
\newblock In {\em Proceedings of the 42nd Meeting of the ACL\/}, pages
  104--111. Barcelona, Spain.

\bibitem[Clark and Curran(2007{\natexlab{b}})]{clark-curran-2007-wide}
Clark, Stephen and James~R. Curran. 2007{\natexlab{b}}.
\newblock Wide-coverage efficient statistical parsing with {CCG} and log-linear
  models.
\newblock {\em Computational Linguistics\/} 33(4):493--552.

\bibitem[Clark et~al.(2015)Clark, Foong, Bulat, and Xu]{java_candc}
Clark, Stephen, Darren Foong, Luana Bulat, and Wenduan Xu. 2015.
\newblock The {J}ava version of the {C\&C} parser: Version 0.95.
\newblock Tech. rep., The University of Cambridge Computer Laboratory,
  Cambridge, UK.

\bibitem[Clark and Hockenmaier(2002)]{clark:lrec02}
Clark, Stephen and Julia Hockenmaier. 2002.
\newblock Evaluating a wide-coverage {CCG} parser.
\newblock In {\em Proceedings of the LREC Beyond Parseval Workshop\/}. Las
  Palmas, Spain.

\bibitem[Clark et~al.(2002)Clark, Hockenmaier, and Steedman]{clark:acl02}
Clark, Stephen, Julia Hockenmaier, and Mark Steedman. 2002.
\newblock Building deep dependency structures with a wide-coverage {C}{C}{G}
  parser.
\newblock In {\em Proceedings of the 40th Meeting of the ACL\/}, pages
  327--334. Philadelphia, PA.

\bibitem[Coecke et~al.(2010)Coecke, Sadrzadeh, and Clark]{coecke:2010}
Coecke, Bob, Mehrnoosh Sadrzadeh, and Stephen Clark. 2010.
\newblock Mathematical foundations for a compositional distributional model of
  meaning.
\newblock In van Bentham and Moortgat, eds., {\em Linguistic Analysis 36 (1-4):
  A Festschrift for Joachim Lambek\/}.

\bibitem[Collins(1997)]{collins:97}
Collins, Michael. 1997.
\newblock Three generative, lexicalised models for statistical parsing.
\newblock In {\em Proceedings of the 35th Meeting of the ACL\/}, pages 16--23.
  Madrid, Spain.

\bibitem[Collins and Brooks(1995)]{collins-brooks-1995-prepositional}
Collins, Michael and James Brooks. 1995.
\newblock Prepositional phrase attachment through a backed-off model.
\newblock In {\em Third Workshop on Very Large Corpora\/}.

\bibitem[Collins and Roark(2004)]{collins_roark:acl04}
Collins, Michael and Brian Roark. 2004.
\newblock Incremental parsing with the perceptron algorithm.
\newblock In {\em Proceedings of the 42nd Meeting of the ACL\/}, pages
  111--118. Barcelona, Spain.

\bibitem[Curran et~al.(2007)Curran, Clark, and
  Bos]{curran-etal-2007-linguistically}
Curran, James, Stephen Clark, and Johan Bos. 2007.
\newblock Linguistically motivated large-scale {NLP} with {C}{\&}{C} and boxer.
\newblock In {\em Proceedings of the 45th Annual Meeting of the Association for
  Computational Linguistics Companion Volume Proceedings of the Demo and Poster
  Sessions\/}, pages 33--36. Prague, Czech Republic: Association for
  Computational Linguistics.

\bibitem[Curran and Clark(2003)]{curran-clark-2003-investigating}
Curran, James~R. and Stephen Clark. 2003.
\newblock Investigating {GIS} and smoothing for maximum entropy taggers.
\newblock In {\em 10th Conference of the {E}uropean Chapter of the Association
  for Computational Linguistics\/}. Budapest, Hungary: Association for
  Computational Linguistics.

\bibitem[Curry and Feys(1958)]{curry:1958}
Curry, Haskell~B. and Robert Feys. 1958.
\newblock {\em Combinatory Logic: Vol. I\/}.
\newblock North Holland: Amsterdam.

\bibitem[Devlin et~al.(2019)Devlin, Chang, Lee, and
  Toutanova]{devlin-etal-2019-bert}
Devlin, Jacob, Ming-Wei Chang, Kenton Lee, and Kristina Toutanova. 2019.
\newblock {BERT}: Pre-training of deep bidirectional transformers for language
  understanding.
\newblock In {\em Proceedings of the 2019 Conference of the North {A}merican
  Chapter of the Association for Computational Linguistics: Human Language
  Technologies, Volume 1 (Long and Short Papers)\/}, pages 4171--4186.
  Minneapolis, Minnesota: Association for Computational Linguistics.

\bibitem[Fowler and Penn(2010)]{fowler-penn-2010-accurate}
Fowler, Timothy A.~D. and Gerald Penn. 2010.
\newblock Accurate context-free parsing with {C}ombinatory {C}ategorial
  {G}rammar.
\newblock In {\em Proceedings of the 48th Annual Meeting of the Association for
  Computational Linguistics\/}, pages 335--344. Uppsala, Sweden: Association
  for Computational Linguistics.

\bibitem[Goodfellow et~al.(2016)Goodfellow, Bengio, and
  Courville]{Goodfellow-et-al-2016}
Goodfellow, Ian, Yoshua Bengio, and Aaron Courville. 2016.
\newblock {\em Deep Learning\/}.
\newblock MIT Press.

\bibitem[Hockenmaier and Steedman(2002)]{hock:acl02}
Hockenmaier, Julia and Mark Steedman. 2002.
\newblock Generative models for statistical parsing with {C}ombinatory
  {C}ategorial {G}rammar.
\newblock In {\em Proceedings of the 40th Meeting of the ACL\/}, pages
  335--342. Philadelphia, PA.

\bibitem[Hockenmaier and Steedman(2007)]{hockenmaier-steedman-2007-ccgbank}
Hockenmaier, Julia and Mark Steedman. 2007.
\newblock {CCG}bank: A corpus of {CCG} derivations and dependency structures
  extracted from the {P}enn {T}reebank.
\newblock {\em Computational Linguistics\/} 33(3):355--396.

\bibitem[Huang et~al.(2012)Huang, Fayong, and Guo]{huang-etal-2012-structured}
Huang, Liang, Suphan Fayong, and Yang Guo. 2012.
\newblock Structured perceptron with inexact search.
\newblock In {\em Proceedings of the 2012 Conference of the North {A}merican
  Chapter of the Association for Computational Linguistics: Human Language
  Technologies\/}, pages 142--151. Montr{\'e}al, Canada: Association for
  Computational Linguistics.

\bibitem[Joshi(1987)]{joshi:87}
Joshi, A.~K. 1987.
\newblock An introduction to tree adjoining grammars.
\newblock In A.~Manaster-Ramer, ed., {\em Mathematics of Language\/}. John
  Benjamins.

\bibitem[Joshi et~al.(1991)Joshi, Vijay-Shanker, and Weir]{joshi:91}
Joshi, A.~K., K.~Vijay-Shanker, and D.~J. Weir. 1991.
\newblock The convergence of mildly context-sensitive grammar formalisms.
\newblock In P.~Sells, S.~Shieber, and T.~Wasow, eds., {\em Foundational Issues
  in Natural Language Processing\/}, pages 31--81. MIT Press, Cambridge MA.

\bibitem[Kasai et~al.(2018)Kasai, Frank, Xu, Merrill, and
  Rambow]{kasai-etal-2018-end}
Kasai, Jungo, Robert Frank, Pauli Xu, William Merrill, and Owen Rambow. 2018.
\newblock End-to-end graph-based {TAG} parsing with neural networks.
\newblock In {\em Proceedings of the 2018 Conference of the North {A}merican
  Chapter of the Association for Computational Linguistics: Human Language
  Technologies, Volume 1 (Long Papers)\/}, pages 1181--1194. New Orleans,
  Louisiana: Association for Computational Linguistics.

\bibitem[Kato and Matsubara(2021)]{kato:emnlp21}
Kato, Yoshihide and Shigeki Matsubara. 2021.
\newblock A new representation for span-based {CCG} parsing.
\newblock In {\em Proceedings of the 2021 Conference on Empirical Methods in
  Natural Language Processing ({EMNLP})\/}. Punta Cana, Dominican Republic:
  Association for Computational Linguistics.

\bibitem[Kitaev et~al.(2019)Kitaev, Cao, and
  Klein]{kitaev-etal-2019-multilingual}
Kitaev, Nikita, Steven Cao, and Dan Klein. 2019.
\newblock Multilingual constituency parsing with self-attention and
  pre-training.
\newblock In {\em Proceedings of the 57th Annual Meeting of the Association for
  Computational Linguistics\/}, pages 3499--3505. Florence, Italy: Association
  for Computational Linguistics.

\bibitem[Kitaev and Klein(2018)]{kitaev-klein-2018-constituency}
Kitaev, Nikita and Dan Klein. 2018.
\newblock Constituency parsing with a self-attentive encoder.
\newblock In {\em Proceedings of the 56th Annual Meeting of the Association for
  Computational Linguistics (Volume 1: Long Papers)\/}, pages 2676--2686.
  Melbourne, Australia: Association for Computational Linguistics.

\bibitem[Kuhlmann et~al.(2015)Kuhlmann, Koller, and
  Satta]{kuhlmann-etal-2015-lexicalization}
Kuhlmann, Marco, Alexander Koller, and Giorgio Satta. 2015.
\newblock Lexicalization and generative power in {CCG}.
\newblock {\em Computational Linguistics\/} 41(2):187--219.

\bibitem[Kuhlmann et~al.(2018)Kuhlmann, Satta, and
  Jonsson]{kuhlmann-etal-2018-complexity}
Kuhlmann, Marco, Giorgio Satta, and Peter Jonsson. 2018.
\newblock On the complexity of {CCG} parsing.
\newblock {\em Computational Linguistics\/} 44(3):447--482.

\bibitem[Kummerfeld et~al.(2010)Kummerfeld, Roesner, Dawborn, Haggerty, Curran,
  and Clark]{kummerfeld-etal-2010-faster}
Kummerfeld, Jonathan~K., Jessika Roesner, Tim Dawborn, James Haggerty, James~R.
  Curran, and Stephen Clark. 2010.
\newblock Faster parsing by supertagger adaptation.
\newblock In {\em Proceedings of the 48th Annual Meeting of the Association for
  Computational Linguistics\/}, pages 345--355. Uppsala, Sweden: Association
  for Computational Linguistics.

\bibitem[Lafferty et~al.(2001)Lafferty, McCallum, and Pereira]{laffertyCrf}
Lafferty, John~D., Andrew McCallum, and Fernando C.~N. Pereira. 2001.
\newblock Conditional random fields: Probabilistic models for segmenting and
  labeling sequence data.
\newblock In {\em Proceedings of the Eighteenth International Conference on
  Machine Learning\/}, ICML '01, pages 282--289. San Francisco, CA, USA: Morgan
  Kaufmann Publishers Inc.
\newblock ISBN 1-55860-778-1.

\bibitem[Lambek(2008)]{lambek:08}
Lambek, Joachim. 2008.
\newblock {\em From Word to Sentence\/}.
\newblock Polimetrica.

\bibitem[Lappin(2021)]{lappin:21}
Lappin, Shalom. 2021.
\newblock {\em Deep Learning and Linguistic Representation\/}.
\newblock Oxford: CRC Press, Taylor and Francis.

\bibitem[Lee et~al.(2016)Lee, Lewis, and Zettlemoyer]{lee-etal-2016-global}
Lee, Kenton, Mike Lewis, and Luke Zettlemoyer. 2016.
\newblock Global neural {CCG} parsing with optimality guarantees.
\newblock In {\em Proceedings of the 2016 Conference on Empirical Methods in
  Natural Language Processing\/}, pages 2366--2376. Austin, Texas: Association
  for Computational Linguistics.

\bibitem[Lewis et~al.(2016)Lewis, Lee, and Zettlemoyer]{lewis-etal-2016-lstm}
Lewis, Mike, Kenton Lee, and Luke Zettlemoyer. 2016.
\newblock {LSTM} {CCG} parsing.
\newblock In {\em Proceedings of the 2016 Conference of the North {A}merican
  Chapter of the Association for Computational Linguistics: Human Language
  Technologies\/}, pages 221--231. San Diego, California: Association for
  Computational Linguistics.

\bibitem[Liu et~al.(2021)Liu, Cohen, Lapata, and Bos]{bos:21}
Liu, Jiangming, Shay~B. Cohen, Mirella Lapata, and Johan Bos. 2021.
\newblock {Universal Discourse Representation Structure Parsing}.
\newblock {\em Computational Linguistics\/} 47(2):445--476.

\bibitem[Lorenz et~al.(2021)Lorenz, Pearson, Meichanetzidis, Kartsaklis, and
  Coecke]{lorenz:21}
Lorenz, Robin, Anna Pearson, Konstantinos Meichanetzidis, Dimitri Kartsaklis,
  and Bob Coecke. 2021.
\newblock {QNLP} in practice: Running compositional models of meaning on a
  quantum computer.
\newblock Tech. Rep. arXiv 2102.12846v1, Cambridge Quantum Computing.

\bibitem[Marcus et~al.(1993)Marcus, Santorini, and Marcinkiewicz]{marcus:93}
Marcus, Mitchell, Beatrice Santorini, and Mary Marcinkiewicz. 1993.
\newblock Building a large annotated corpus of {E}nglish: The {P}enn
  {T}reebank.
\newblock {\em Computational Linguistics\/} 19(2):313--330.

\bibitem[Miyao and Tsujii(2008)]{miyao-tsujii-2008-feature}
Miyao, Yusuke and Jun{'}ichi Tsujii. 2008.
\newblock Feature forest models for probabilistic {HPSG} parsing.
\newblock {\em Computational Linguistics\/} 34(1):35--80.

\bibitem[Moortgat(1997)]{moortgat:97}
Moortgat, Michael. 1997.
\newblock Categorial type logics.
\newblock In J.~van Benthem and A.~ter Meulen, eds., {\em Handbook of Logic and
  Language\/}, chap.~2, pages 93--177. Elsevier, Amsterdam and MIT Press,
  Cambridge MA.

\bibitem[Ratnaparkhi(1996)]{ratnaparkhi:96}
Ratnaparkhi, Adwait. 1996.
\newblock A maximum entropy model for part-of-speech tagging.
\newblock In {\em Proceedings of the EMNLP Conference\/}, pages 133--142.
  Philadelphia, PA.

\bibitem[Riezler et~al.(2002)Riezler, King, Kaplan, Crouch, III, and
  Johnson]{riezler:acl02}
Riezler, Stefan, Tracy~H. King, Ronald~M. Kaplan, Richard Crouch, John
  T.~Maxwell III, and Mark Johnson. 2002.
\newblock Parsing the {W}all {S}treet {J}ournal using a {L}exical-{F}unctional
  {G}rammar and discriminative estimation techniques.
\newblock In {\em Proceedings of the 40th Meeting of the ACL\/}, pages
  271--278. Philadelphia, PA.

\bibitem[Rimell and Clark(2008)]{rimell-clark-2008-adapting}
Rimell, Laura and Stephen Clark. 2008.
\newblock Adapting a lexicalized-grammar parser to contrasting domains.
\newblock In {\em Proceedings of the 2008 Conference on Empirical Methods in
  Natural Language Processing\/}, pages 475--484. Honolulu, Hawaii: Association
  for Computational Linguistics.

\bibitem[Rimell et~al.(2009)Rimell, Clark, and
  Steedman]{rimell-etal-2009-unbounded}
Rimell, Laura, Stephen Clark, and Mark Steedman. 2009.
\newblock Unbounded dependency recovery for parser evaluation.
\newblock In {\em Proceedings of the 2009 Conference on Empirical Methods in
  Natural Language Processing\/}, pages 813--821. Singapore: Association for
  Computational Linguistics.

\bibitem[Shieber(1985)]{shieber:85}
Shieber, Stuart. 1985.
\newblock Evidence against the context-freeness of natural language.
\newblock {\em Linguistics and Philosophy\/} 8:333--343.

\bibitem[Stanojevi{\'c} and Steedman(2019)]{stanojevic-steedman-2019-ccg}
Stanojevi{\'c}, Milo{\v{s}} and Mark Steedman. 2019.
\newblock {CCG} parsing algorithm with incremental tree rotation.
\newblock In {\em Proceedings of the 2019 Conference of the North {A}merican
  Chapter of the Association for Computational Linguistics: Human Language
  Technologies, Volume 1 (Long and Short Papers)\/}, pages 228--239.
  Minneapolis, Minnesota: Association for Computational Linguistics.

\bibitem[Stanojevi{\'c} and Steedman(2020)]{stanojevic-steedman-2020-max}
Stanojevi{\'c}, Milo{\v{s}} and Mark Steedman. 2020.
\newblock Max-margin incremental {CCG} parsing.
\newblock In {\em Proceedings of the 58th Annual Meeting of the Association for
  Computational Linguistics\/}, pages 4111--4122. Online: Association for
  Computational Linguistics.

\bibitem[Steedman(1996)]{steedman:96}
Steedman, Mark. 1996.
\newblock {\em Surface Structure and Interpretation\/}.
\newblock Cambridge, MA: The MIT Press.

\bibitem[Steedman(2000)]{steedman:2000}
Steedman, Mark. 2000.
\newblock {\em The Syntactic Process\/}.
\newblock Cambridge, MA: The MIT Press.

\bibitem[Stern et~al.(2017)Stern, Andreas, and Klein]{stern-etal-2017-minimal}
Stern, Mitchell, Jacob Andreas, and Dan Klein. 2017.
\newblock A minimal span-based neural constituency parser.
\newblock In {\em Proceedings of the 55th Annual Meeting of the Association for
  Computational Linguistics (Volume 1: Long Papers)\/}, pages 818--827.
  Vancouver, Canada: Association for Computational Linguistics.

\bibitem[Tian et~al.(2020)Tian, Song, and Xia]{tian-etal-2020-supertagging}
Tian, Yuanhe, Yan Song, and Fei Xia. 2020.
\newblock Supertagging {C}ombinatory {C}ategorial {G}rammar with attentive
  graph convolutional networks.
\newblock In {\em Proceedings of the 2020 Conference on Empirical Methods in
  Natural Language Processing (EMNLP)\/}, pages 6037--6044. Online: Association
  for Computational Linguistics.

\bibitem[Vaswani et~al.(2016)Vaswani, Bisk, Sagae, and
  Musa]{vaswani-etal-2016-supertagging}
Vaswani, Ashish, Yonatan Bisk, Kenji Sagae, and Ryan Musa. 2016.
\newblock Supertagging with {LSTM}s.
\newblock In {\em Proceedings of the 2016 Conference of the North {A}merican
  Chapter of the Association for Computational Linguistics: Human Language
  Technologies\/}, pages 232--237. San Diego, California: Association for
  Computational Linguistics.

\bibitem[Vaswani et~al.(2017)Vaswani, Shazeer, Parmar, Uszkoreit, Jones, Gomez,
  Kaiser, and Polosukhin]{vaswani2017attention}
Vaswani, Ashish, Noam Shazeer, Niki Parmar, Jakob Uszkoreit, Llion Jones,
  Aidan~N Gomez, {\L}ukasz Kaiser, and Illia Polosukhin. 2017.
\newblock Attention is all you need.
\newblock In {\em Advances in Neural Information Processing Systems\/}, pages
  5998--6008.

\bibitem[Vijay-Shanker and Weir(1993)]{vijay:93}
Vijay-Shanker, K. and David Weir. 1993.
\newblock Parsing some constrained grammar formalisms.
\newblock {\em Computational Linguistics\/} 19:591--636.

\bibitem[Wang and Chang(2016)]{wang-chang-2016-graph}
Wang, Wenhui and Baobao Chang. 2016.
\newblock Graph-based dependency parsing with bidirectional {LSTM}.
\newblock In {\em Proceedings of the 54th Annual Meeting of the Association for
  Computational Linguistics (Volume 1: Long Papers)\/}, pages 2306--2315.
  Berlin, Germany: Association for Computational Linguistics.

\bibitem[Weir(1992)]{weir:92}
Weir, David. 1992.
\newblock A geometric hierarchy beyond context-free languages.
\newblock {\em Theoretical Computer Science\/} 104(2):235--261.

\bibitem[Xu(2016)]{xu-2016-lstm}
Xu, Wenduan. 2016.
\newblock {LSTM} shift-reduce {CCG} parsing.
\newblock In {\em Proceedings of the 2016 Conference on Empirical Methods in
  Natural Language Processing\/}, pages 1754--1764. Austin, Texas: Association
  for Computational Linguistics.

\bibitem[Xu et~al.(2015)Xu, Auli, and Clark]{xu-etal-2015-ccg}
Xu, Wenduan, Michael Auli, and Stephen Clark. 2015.
\newblock {CCG} supertagging with a recurrent neural network.
\newblock In {\em Proceedings of the 53rd Annual Meeting of the Association for
  Computational Linguistics and the 7th International Joint Conference on
  Natural Language Processing (Volume 2: Short Papers)\/}, pages 250--255.
  Beijing, China: Association for Computational Linguistics.

\bibitem[Xu et~al.(2016)Xu, Auli, and Clark]{xu-etal-2016-expected}
Xu, Wenduan, Michael Auli, and Stephen Clark. 2016.
\newblock Expected {F}-measure training for shift-reduce parsing with recurrent
  neural networks.
\newblock In {\em Proceedings of the 2016 Conference of the North {A}merican
  Chapter of the Association for Computational Linguistics: Human Language
  Technologies\/}, pages 210--220. San Diego, California: Association for
  Computational Linguistics.

\bibitem[Xu et~al.(2014)Xu, Clark, and Zhang]{xu-etal-2014-shift}
Xu, Wenduan, Stephen Clark, and Yue Zhang. 2014.
\newblock Shift-reduce {CCG} parsing with a dependency model.
\newblock In {\em Proceedings of the 52nd Annual Meeting of the Association for
  Computational Linguistics (Volume 1: Long Papers)\/}, pages 218--227.
  Baltimore, Maryland: Association for Computational Linguistics.

\bibitem[Yeung and Kartsaklis(2021)]{yeung:21}
Yeung, Richie and Dimitri Kartsaklis. 2021.
\newblock A {CCG}-based version of the {D}is{C}o{C}at framework,.
\newblock In {\em Proceedings of SemSpace 2021\/}.

\bibitem[Yoshikawa et~al.(2017)Yoshikawa, Noji, and
  Matsumoto]{yoshikawa-etal-2017-ccg}
Yoshikawa, Masashi, Hiroshi Noji, and Yuji Matsumoto. 2017.
\newblock {A}* {CCG} parsing with a supertag and dependency factored model.
\newblock In {\em Proceedings of the 55th Annual Meeting of the Association for
  Computational Linguistics (Volume 1: Long Papers)\/}, pages 277--287.
  Vancouver, Canada: Association for Computational Linguistics.

\bibitem[Zettlemoyer and Collins(2005)]{zettlemoyer:05}
Zettlemoyer, Luke~S. and Michael Collins. 2005.
\newblock Learning to map sentences to logical form: Structured classification
  with probabilistic categorial grammars.
\newblock In {\em Proceedings of the 21st Conference on Uncertainty in
  Artificial Intelligence\/}. Edinburgh, UK.

\bibitem[Zhang and Clark(2011)]{zhang-clark-2011-shift}
Zhang, Yue and Stephen Clark. 2011.
\newblock Shift-reduce {CCG} parsing.
\newblock In {\em Proceedings of the 49th Annual Meeting of the Association for
  Computational Linguistics: Human Language Technologies\/}, pages 683--692.
  Portland, Oregon, USA: Association for Computational Linguistics.

\end{thebibliography}

\end{document}